\def\year{2022}\relax
\documentclass[letterpaper]{article} % DO NOT CHANGE THIS

\usepackage{amsmath}
\usepackage{amsfonts}
\usepackage{kotex}
\usepackage{comment}
\usepackage{subfig}

\usepackage{xcolor}

\usepackage{subfig}
\usepackage{booktabs}
\usepackage{multirow}
\usepackage{aaai22}  % DO NOT CHANGE THIS
\usepackage{times}  % DO NOT CHANGE THIS
\usepackage{helvet}  % DO NOT CHANGE THIS
\usepackage{courier}  % DO NOT CHANGE THIS
\usepackage[hyphens]{url}  % DO NOT CHANGE THIS
\usepackage{graphicx} % DO NOT CHANGE THIS
\usepackage{natbib}  % DO NOT CHANGE THIS AND DO NOT ADD ANY OPTIONS TO IT
\usepackage{caption} % DO NOT CHANGE THIS AND DO NOT ADD ANY OPTIONS TO IT
\DeclareCaptionStyle{ruled}{labelfont=normalfont,labelsep=colon,strut=off} % DO NOT CHANGE THIS
\usepackage{algorithm}
\usepackage{algorithmic}

\usepackage{newfloat}
\usepackage{listings}
\floatstyle{ruled}
\newfloat{listing}{tb}{lst}{}
\floatname{listing}{Listing}
\title{Spatio-Temporal Road Traffic Prediction using Real-time Regional Knowledge}

\author {
Sumin Han,\textsuperscript{\rm 1}
Jisun An,\textsuperscript{\rm 2}
Dongman Lee\textsuperscript{\rm 1}
}

\affiliations {
\textsuperscript{\rm 1} School of Computing, Korea Advanced Institute of Science and Technology\\
\textsuperscript{\rm 2} Luddy School of Informatics, Computing, and Engineering, Indiana University Bloomington\\
hsm6911@kaist.ac.kr, jisunan@iu.edu, dlee@kaist.ac.kr
}
\usepackage{bibentry}
\begin{document}

\maketitle

\begin{abstract}

  For traffic prediction in transportation services such as car-sharing and ride-hailing, mid-term road traffic prediction (within a few hours) is considered essential. However, the existing road-level traffic prediction has mainly studied how significantly micro traffic events propagate to the adjacent roads in terms of short-term prediction. On the other hand, recent attempts have been made to incorporate regional knowledge such as POIs, road characteristics, and real-time social events to help traffic prediction. However, these studies lack in understandings of different modalities of road-level and region-level spatio-temporal correlations and how to combine such knowledge. This paper proposes a novel method that embeds real-time region-level knowledge using POIs, satellite images, and real-time LTE access traces via a regional spatio-temporal module that consists of dynamic convolution and temporal attention, and conducts bipartite spatial transform attention to convert into road-level knowledge. Then the model ingests this embedded knowledge into a road-level attention-based prediction model. Experimental results on real-world road traffic prediction show that our model outperforms the baselines.
\end{abstract}

\section{Introduction}

Traffic prediction refers to predicting taffic-related values such as traffic speed or volume~\cite{yuan2021survey}. From the perspective of urban planning, traffic prediction is essential as it helps to determine traffic signal operation or plan public transportation. From an individual's point of view, it becomes necessary to plan a trip such as when to leave or which transportation to use given the traffic condition. 
Traffic prediction is broadly classified into three categories depending on the time period: short-term (a few minutes to an hour), mid-term (within a few hours), and long-term (more than one day) prediction~\cite{hou2016repeatability}. A typical application of short-term prediction is a real-time navigation system, and mid/long-term predictions are used for transportation planning.
% while that of mid/long-term prediction is transportation planning.
%In particular, mid-term road traffic prediction is considered essential in transportation forecastings such as car-sharing and ride-hailing \textcolor{red}{[refs]}. 

\begin{figure}[h]
    \centering
    \includegraphics[width=8cm]{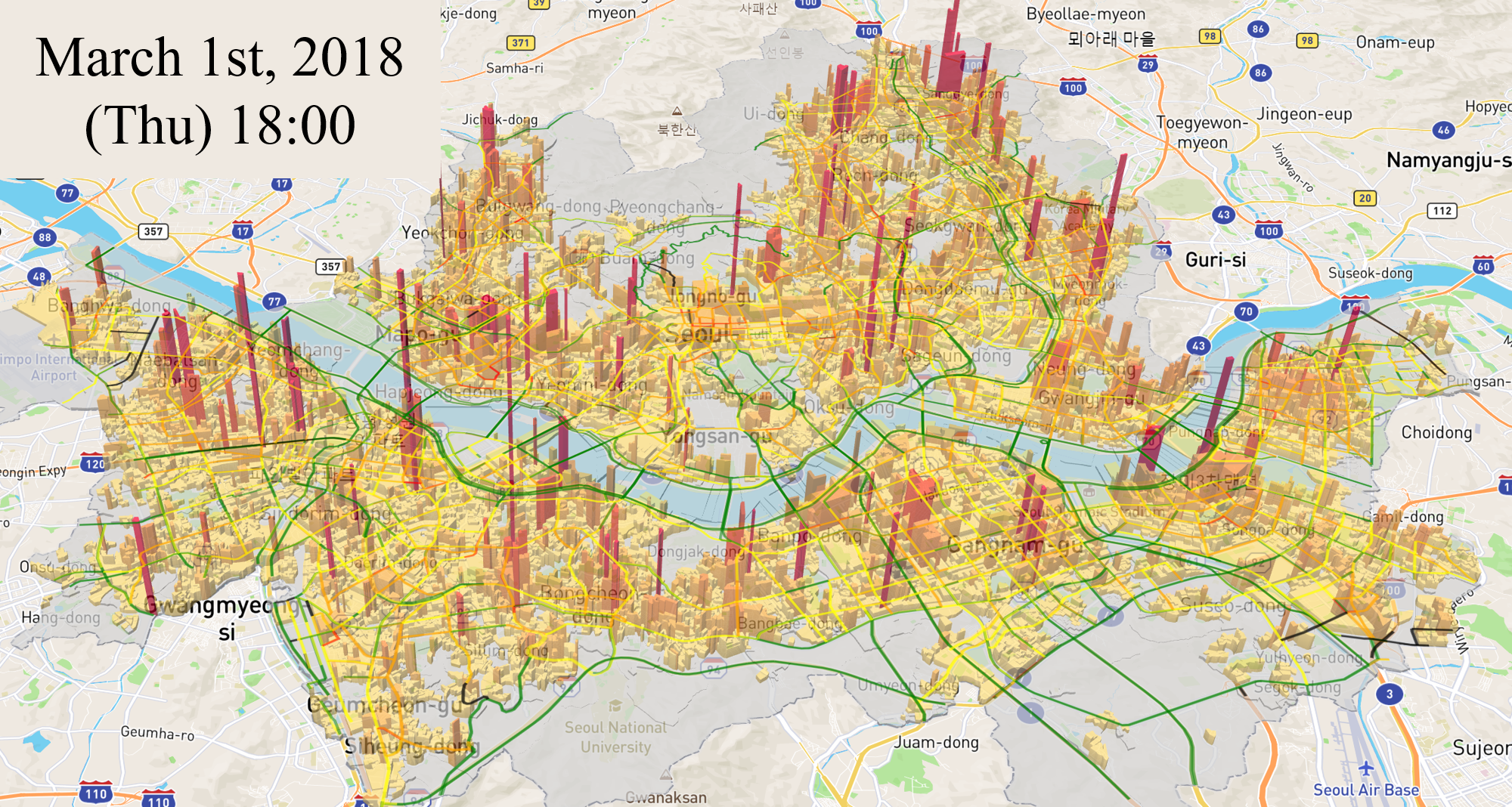}
    \caption{Real-time road traffic speed and regional population (LTE access traces) visualization in Seoul, Korea.} %\todo{no mention of figure 1 in the body?}
    \label{fig:overview}
\end{figure}

While most of the previous deep learning approaches have tackled short-term traffic prediction problems~\cite{jiang2021survey}, the mid/long-term traffic prediction has recently become increasingly important for advanced traffic management systems (ATMS), especially with the rise of autonomous cars and car-sharing services~\cite{lana2018road,hoermann2017probabilistic}.
Unlike short-term prediction, which mainly utilizes graph information of road network, recent studies have shown that mid/long-term traffic prediction benefits from regional knowledge (e.g., Point of Interests (POIs))~\cite{zhang2017impact,hu2018discovering,gonzalez2019transit}.
However, regional knowledge is not static but \textit{dynamic} -- it may change as a social event occurs or travel demand changes by unexpected causes~\cite{huang2019exploration}.
This would be particularly true for those cities with a large population and high density.
Thus, existing models that incorporate static regional knowledge for traffic prediction may fail in a complex urban environment where traffic is affected by real-time regional exploitation.

As live regional datasets are becoming more available such as real-time regional population or ridership data as Figure~\ref{fig:overview}, there is increasing interest to utilize this regional knowledge to produce better traffic prediction.
In incorporating real-time regional knowledge for traffic prediction, three main challenges remain to be solved as follows: 

\noindent\textit{1) Incorporating region-level and road-level data equally:}
%Utilizing regional information \textbf{without} modality transformation at the data processing phase:
%transform either graph to region or region to graph at the data processing phase\cite{zhang2020curb,lv2020temporal,yuan2018hetero}. However, they do not incorporate individual module that
Previous work~\cite{zhang2020curb,lv2020temporal,yuan2018hetero} did not handle regional-level and road-level data separately, which may cause performance degradation.
For example, it would produce a better result if regional data is analyzed using convolutional neural network (CNN) while road data is incorporated using graph convolutional neural network (GCN).

\noindent\textit{2) Dynamic regional knowledge learning:} 
Although a model is given real-time regional traffic data and finds certain patterns, it can not solely explain why such patterns occur without background information about the region. The model can have limited learning performance if there is no reasoning from the built-in environment to analyze the regional data.
% Although a model is utilizing real-time regional traffic data, it can have limited learning performance without background information about the region as it can not explain why certain patterns occur. Thus, it needs to have a mechanism for  from the built-in environment to analyze the regional data.
% While real-time regional traffic data can be a significant use for traffic prediction, how to incorporate it to a model is challenging. Without proper reasoning from the built-in environment, such model may have limited learning performance.
\noindent\textit{3) Transformation of regional knowledge into a road:} 
% To determine how much regional information to give for each road for modality transformation, transportation studies typically bound 500m nearby region, which is a walking distance to transit~\cite{van2020exploring}. 
For modality transformation, in determining how much regional information to use for each road, transportation studies typically bound 500m nearby region, which is a walking distance to transit~\cite{van2020exploring}.
However, this bound should be adapted differently depending on the road characteristics.
% However, this bound should be differently adapted for roads depending on their characteristics.
%Although a model is given real-time regional traffic data, the model should be able to reason from the built-in environment of the region to analyze the pattern.

In this study, we propose a novel method to embed both real-time and static regional knowledge for traffic prediction. In particular, we use fine-grained, hourly population estimated from LTE access trace counts together with POIs and satellite images to capture the built-in environment.
Our model is composed of a regional spatio-temporal module that consists of dynamic convolution and temporal attention, and conducts bipartite spatial transform attention to convert into road-level knowledge. Then the model ingests this embedded knowledge into a road-level attention-based prediction model. 
Experimental results show that our model outperforms the baselines. 
We summarize our contributions as follows:

\begin{itemize}
    \item To the best of our knowledge, this is the first research that incorporates real-time regional population data for road-level traffic prediction with each modality training. For this, we propose a novel method that learns real-time region-level knowledge for road-level prediction. 
    \item We propose a dynamic convolution based on regional correlation and distance, and a bipartite masked transform attention which adds a gaussian mask to train attention scores from the nearby region for each road.
    \item We construct real-world dataset in Seoul, Korea, and the evaluation shows that our model outperforms by utilizing real-time regional knowledge. We make our dataset publicly available to contribute research community.%\footnote{We blind the public pointer during the review process, but we provide datasets in the supplementary material.}
\end{itemize}

\section{Related Work}
\subsection{Road Traffic Prediction}

The biggest challenge of the road traffic prediction model is to use the road connectivity network for training, as the impact of connection in a road graph is higher than the Euclidean distance between traffic sensors. 
DCRNN~\cite{li2017diffusion} captures spatio information of road graph through diffusion convolution and combines with RNN module to learn spatio-temporal information.
%STGCN?
ST-MetaNet~\cite{pan2019urban_stmetanet} learns the meta knowledge of nodes and edges, and extracts the weights for an RNN cell from meta information to give illusion that each traffic sensor has its own RNN cell that conducts temporal prediction. 
GMAN~\cite{zheng2020gman} leverages spatio-temporal embeddings by Node2Vec~\cite{grover2016node2vec} and timestamps, and applies a spatial attention and a temporal attention combined with a gated fusion. 
ST-GRAT~\cite{park2020st} proposes sentinel mechanism that complement a spatial attention when there are nodes have less spatial correlation.

\subsection{Regional Traffic Prediction}

For regional traffic prediction models, it is important to learn features for each adjacent cell in a 2D grid. Therefore, a model that passes spatial information through regional filters like convolutional neural network (CNN) and learns temporal information through RNN/temporal attention is mostly proposed.
ConvLSTM~\cite{xingjian2015convolutional} is an improved structure that can be used for next video frame prediction by combining a convolution layer instead of fully Connected layer inside LSTM.
HeteroConvLSTM~\cite{yuan2018hetero} applies this ConvLSTM structure for regional spatio-temporal traffic accident prediction.
MC-STGCN~\cite{tang2021multi} incorporates spatial correlations among regions using regional community graphs and leverages GRU for temporal correlation learning for better taxi ridership demand prediction.
%ST-MetaNet~\cite{pan2019urban_stmetanet} also test its model on regional traffic data of taxi ridership in Beijing using meta data such as regional information (POIs) and pair-wise cell information (e.g., the number of roads and lanes).

\subsection{Multi-modal Traffic Prediction}
There are a handful of research that use both real-time region-level and road-level data for prediction. 
%They mainly transform road-level traffic data into region-level to enhance road traffic prediction.
CurbGAN~\cite{zhang2020curb} proposes generative adversarial network (GAN) that estimates regional traffic speed from travel demand inferred from taxi ridership.
%It leverages dynamic convolution that is enhanced than CNN by creating a regionally optimized convolutional filter for better spatial knowledge learning.
HeteroConvLSTM~\cite{yuan2018hetero} transforms the road graph onto 2D-grid region to embed road information for regional traffic prediction. 
On the other hand, there are several trials to use regional features for traffic prediction. 
\citet{lv2020temporal} utilizes POIs with gaussian kernel from each road to capture road-level geographical information to enhance road traffic prediction. 
\citet{lin2020passenger_range} captures land-use proportion within 500m (walking distance) of a subway station for station-level traffic prediction.

\section{Preliminaries}

We define our problem as a road traffic prediction using both road and regional traffic history.
%For a timestamp $t$, we denote road-level traffic data (e.g. traffic speed, traffic flow volume) as  $X^{(t)} \in \mathbb{R}^{N_X}$, where $N_X$ is the number of road traffic sensors.
We denote road traffic data (e.g. traffic speed, traffic flow volume) as $X$,
%We denote region-level traffic data (e.g. population density, travel demand) as $Z^{(t)} \in \mathbb{R}^{N_Z}$, where $N_Z = N_{h} \times  N_{w}$ is the number of regional grid cells of $N_{h}$ height and $N_{w}$ width.
and regional traffic data (e.g. population density, travel demand) as $Z$.
For a timestamp $t$, $Z^{(t)}\in \mathbb{R}^{N_Z}$, where $N_Z = N_h \times N_w$ is the number of grid cells of a $N_h$-height and $N_w$-width rectangular region, and $X^{(t)} \in \mathbb{R}^{N_X}$ where $N_X$ is the number of traffic sensors of the roads.
%본 논문에서는 크게 road-level과 region-level의 multi-modal 시공간 교통 데이터를 사용한다. 먼저 road-level spatio-temporal traffic 값(e.g. 속도, 교통량)을 $X$로 표기하며, $N_X$를 도로 네트워크에서 교통량을 기록하는 센서의 개수라고 할 때 임의의 timestamp $t$에 대하여 $X^{(t)} \in \mathbb{R}^{N_X}$ 이다. 다음 region-level spatio-temporal traffic 값(e.g. population density, travel demand)를 $Z$로 표기하며, 활용 데이터가 커버하는 지역의 2D-grid cell unit의 가로를 $N_w$, 세로를 $N_h$라고 할 때 cell의 전체개수는 $N_Z = N_w \times N_h$이며, 임의의 timestamp $t$ 에 대하여 $Z^{(t)} \in \mathbb{R}^{N_Z}$ 이다.
The problem is formulated as finding an optimal function $h(\cdot)$ that inputs $P$-temporal history of road and regional traffic data,
%($[X^{(t-P+1)}, ..., X^{(t)}] \in \mathbb{R}^{N_X \times P}$ and $[Z^{(t-P+1)}, ..., Z^{(t)}] \in \mathbb{R}^{N_Z \times P}$, respectively)
to output $Q$-sequence road traffic prediction of a timestamp $t$, as Equation~\ref{eq:problem_def}.
%($[\hat{X}^{(t+1)}, ..., \hat{X}^{(t+Q)}] \in \mathbb{R}^{N_X \times Q}$)
%as described in Equation \ref{eq:problem_def}.

\begin{multline}
    [X^{(t-P+1)}, ..., X^{(t)}, Z^{(t-P+1)}, ..., Z^{(t)}]  \\
    \xrightarrow{h(\cdot)} [\hat{X}^{(t+1)}, ..., \hat{X}^{(t+Q)}]
\label{eq:problem_def}
\end{multline}

% \subsection{Activation Function}

\subsection{Multi-head Attention Mechanism}
We leverage dot-product attention~\cite{vaswani2017attention} in our method, which has the following form:

\begin{equation}\label{eq:attention}
        H = \text{Att}(Q, K, V) = \text{S}(Q, K) V = \text{softmax}(\frac{QK^T}{\sqrt{d_h}})V
\end{equation}

% \begin{equation}
%     \begin{gathered}\label{eq:attention}
%         H = Attention(Q, K, V) = Score(Q, K) V ,  \\
%         Score(Q, K) = softmax(\frac{QK^T}{\sqrt{d_h}})
%     \end{gathered}
% \end{equation}

with $\text{Att}(\bullet)$ is the attention function, and $Q$ (query), $K$ (key), $V$ (value) is formulated by

\begin{equation}\label{eq:attention}
    Q=f_1 (X_q ), \ \  K = f_2 (X_k ) , \ \ V = f_3 (X_v )
\end{equation}

where $X_q \in \mathbb{R} ^ {N_Q \times d_q}$, $X_k \in \mathbb{R} ^ {N_P \times d_k}$, $X_v \in \mathbb{R} ^ {N_P \times d_v}$ are inputs for $Q$, $K$, $V$ and $f_1$, $f_2$, $f_3$ are activation functions.
In this paper, we denote a non-linear activation function as 
% \begin{equation}\label{eq:activation}
%     f(x) = \text{ReLU}(x W + b)
% \end{equation}
$f(x) = \text{ReLU}(x \mathbf{W} + \mathbf{b})$ 
where $\mathbf{W}$, $\mathbf{b}$ are learnable parameters.
$N_Q$ is the number of queries and $N_P$ is the number of keys and values. 
If we use the module as self-attention, $N_Q$ and $N_P$ become equal as we input $X_q$, $X_p$, $X_k$ the same value. 
In this work, we employ $f_1$, $f_2$, $f_3$ to produce the embedded dimensions of $Q$, $K$, $V$ to be equally $d_h$. 
The attention score $\text{S} (Q, K)$ calculates $N_Q \times N_P$ values of how  $V \in \mathbb{R}^{N_P \times d_h}$ will be summed. 
%Note that, each row of attention score $\text{S} (Q, K) _ {[q_i, :]}$ is summed to 1, as it uses softmax. 
In multi-headed attention, we concatenate (denoted as $\Vert$) the output of $\text{K}$-attention heads, and apply an activation $f_o$.
In this study, we set $K\times d_h$ to be $D$ for all attention mechanism.

\begin{equation}\label{eq:multihead}
    \text{MHAtt}(Q, K, V) = f_o ( \Vert _{k=1}^{k=\text{K}} \{ H_k \})
\end{equation}

We also denote self attention as  $\text{SelfAtt}(X) = \text{MHAtt}(f_1(X), f_2(X), f_3(X))$ for an illustration purpose.
% \begin{equation}\label{eq:selfattn}
% \end{equation}

\subsection{Masked Attention Networks}

%In this study, we use mask attention networks for self-attention and transform attention.

We leverage masked attention mechanism used in~\citet{sperber2018self}. The idea is to add mask $M$ to attention scores before $\text{softmax}$ is applied.

% \begin{equation}\label{eq:maskattn}
% \begin{gather}
    
% \end{gather}
%         \text{Att}^M(\bullet) = \text{S}_i^M(Q, K, M) V_i = \text{softmax}(\frac{QK^T}{\sqrt{d_h}} + M)V
% \end{equation}

\begin{equation}\label{eq:maskattn}
\begin{gathered}
    \text{Att}^M(Q, K, V, M) = \text{S}^M(Q, K, M) V \ \ \  \text{where} \\
    \text{S}^M(Q, K, M)  = \text{softmax}(\frac{QK^T}{\sqrt{d_h}} + M)
\end{gathered}
\end{equation}

In general, we can apply strict masking $M \in \{0, -\infty \}$ to take attention on specific values. % or not based on known information.
%(e.g. edge connection between roads). 
We can also apply different masks for multi-head attention heads. We denote multi-head masked attention networks as $\text{MHAtt}^M(\bullet)$.

\section{Methodology}

\begin{figure}[h]
    \includegraphics[width=8.3cm]{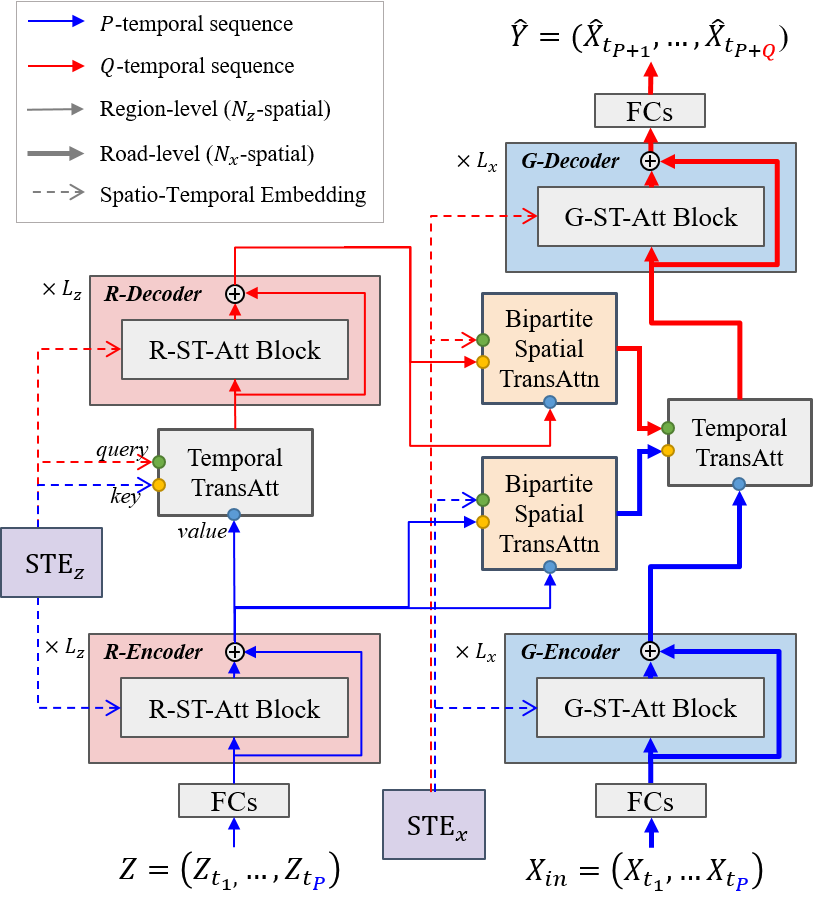}
    \caption{Proposed model architecture. Different arrow types represents different spatial- ($N_Z$: thin, $N_X$: thick), temporal-dimension ($P$: blue, $Q$: red), or spatio-temporal embedding ($\text{STE}_X$, $\text{STE}_Z$: dashed). % as explained in the left-top textbox. 
    %Blue represents P-temporal dimension, and red represents Q-temporal dimension. Arrows represents road-level spatio dimensional ($N_X$) is thicker than region-level dimensional ($N_Z$) features. Spatio-temporal embedding is represented by dashed lines. 
    For each transform attention block (Temporal TransAttn, Bipartite Spatial TransAttn), query and key are given from the left side (green and yellow bullets) and value is given from the bottom side (a blue bullet).
    }
    \label{fig:method}
\end{figure}

Figure~\ref{fig:method} shows overall architecture of our proposed model. Our approach is intuitioned by GMAN~\cite{zheng2020gman} that leverages spatio-temporal embeddings for self-attention and transform attention. 
The biggest difference of our model to GMAN is that we give regional knowledge as query and key to the temporal transform attention of the GMAN block. 
%\textcolor{red}{[don't we need to give a short summary of our differentiation from GMAN?]}
To begin with, there are spatio-temporal embeddings for road-level ($\text{STE}_X$) and region-level ($\text{STE}_Z$) which mark the spatial information and timestamps.
On the left side, region-level encoder and decoder compute representations of P- and Q-sequence regional data. On the right side, the model leverages the bipartite spatial transform attention to convert this region-level knowledge into road-level knowledge and feed the road-level temporal transform attention. 
A $L$-stacked ST-Att block simply concatenates a spatio-temporal embedding with the previous hidden output as its input and conducts self-attention, while making the residual addition.
A transform attention block (temporal and bipartite spatial) uses different attention query and key depends on how it will transform the input whether in temporal dimension ($P$$ \rightarrow $$Q$) or in spatial dimension ($N_Z$$ \rightarrow $$N_X$).
In our implementation, all modules produce $D$-dimensional outputs for residual computation.% and simplicity.
%We explain more detail of each component in the following subsections.

\subsection{Spatio-Temporal Embedding}

\begin{figure}[h]
\centering
    \includegraphics[width=8.5cm]{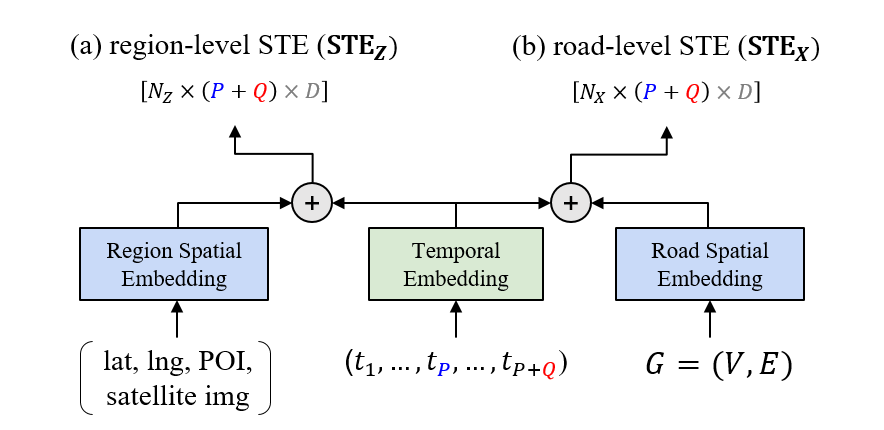}
    \caption{Spatio-temporal embedding (STE) for (a) region-level ($\text{STE}_Z$) and (b) road-level ($\text{STE}_X$) for an attention module. }
    \label{fig:ste}
\end{figure}

In order to train spatio-temporal knowledge based on the road-network and geographic regional environment, we give different spatio-temporal embeddings (STE) for region-level and road-level. % based on different spatial modality.
Figure~\ref{fig:ste} shows how we create spatio-temporal embedding to be used in an attention module.

The STE for regional data ($\text{STE}_Z$) is produced by adding regional spatial embedding with temporal embedding. For a region spatial embedding, we use location (latitude, longitude), POI counts, and satellite image features. We concatenate these geographical features for each cell and apply a 2-layer fully connected network (FCN) to create $D$-dimensional output. For a temporal embedding, we concatenate one-hot encodings of hour-of-day and day-of-week ($\mathbb{R}^{24+7}$) and apply a 2-layer FCN to create an output. Finally, we add these values in combination of $N_Z \times (P+Q)$ to create $\text{STE}_Z \in \mathbb{R}^{N_Z \times (P+Q) \times D}$.

The STE for road data ($\text{STE}_X$) is produced as similar to $\text{STE}_Z$. But, for the road spatial embedding, we use road embedding feature trained from Node2Vec~\cite{grover2016node2vec} using road network, and conduct a 2-layer FCN to create $D$-dimensional output. For the temporal embedding, we share the same one as $\text{STE}_X$.  Then we add these values in combination of $N_X \times (P+Q)$ to create $\text{STE}_X \in \mathbb{R}^{N_X \times (P+Q) \times D}$.

\subsection{Spatio Temporal Attention Block}

\begin{figure}[h]
    \centering
    \includegraphics[width=7.5cm]{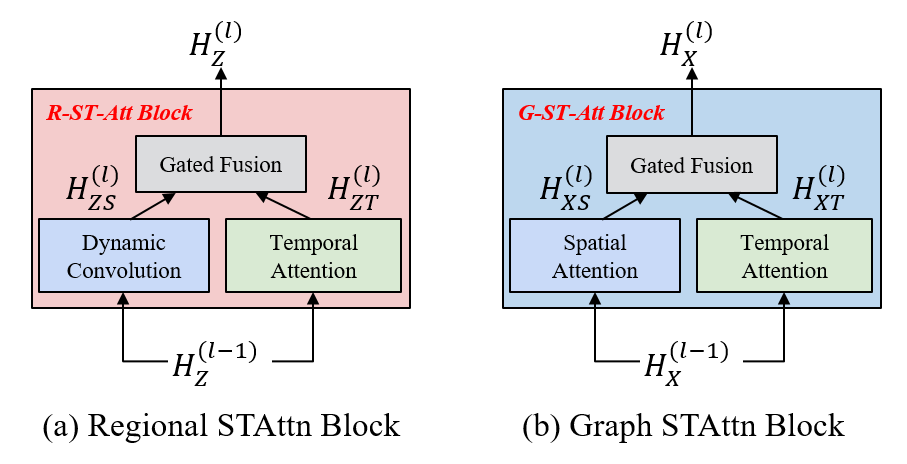}
    \caption{Spatio-temporal attention block for (a) region-level (R-ST-Att Block) and (b) road(graph)-level (G-ST-Att Block) input data. }
    \label{fig:stablock}
\end{figure}

Figure~\ref{fig:stablock} shows the how we implement regional and graph spatio-temporal attention (ST-Att) block. 
A ST-Att block learns spatio-temporal patterns of the input data by self-attention mechanism. The  regional ST-Att (R-ST-Att) block captures spatial pattern by dynamic convolution and temporal pattern by temporal attention. The road graph ST-Att block (G-ST-Att) captures spatial and temporal pattern by respective attention mechanism. 
%by temporal attention. 
In both modules, the gated fusion is applied to sum spatial output and temporal output by trainable ratio that produces optimal output.

\subsubsection{Dynamic Convolution}

% \begin{figure}[h]
%     \centering
%     \includegraphics[width=8cm]{dyconv}
%     \caption{Dynamic convolution}
%     \label{fig:dyconv}
% \end{figure}

A pair of cell is correlated if their regional characteristics are similar, while there is a less correlation between distant cells. 
%Although each regional cell is affected by its nearby regions, they may not be correlated if their built-in environment is too different. 
To give regional correlation degraded by distance, we apply a dynamic graph convolution~\cite{zhang2020curb} with different edge weights based on Pearson correlation of real-time population on total timespan and distance. We formulate it as 
$H^{(l)} = f(\tilde{A}^\text{R} H^{(l-1)}) = \text{ReLU}(\tilde{A}^\text{R} H^{(l-1)} \mathbf{W})$
% \end{equation}
with $\tilde{} $ is a row-normalization, $\mathbf{W} \in \mathbb{R}^{D \times D}$ is a learnable parameter, and $A^{\text{R}} \in \mathbb{R}^{N_Z \times N_Z}$ is an edge weight matrix calculated as 

\begin{equation}
A^\text{R} _{i,j} = \left\{\begin{matrix}
\ r_{i,j} \exp({-(d_{i,j}/\sigma_\text{dist} )^2}) & \text{if} \  \ r_{i,j} > \ \lambda_r \\ % \ \text{and} \ d_{ij} \leq \lambda_d \\ 
0 & \text{otherwise}
\end{matrix}\right.
\end{equation}

where 
$r_{i,j}$ is a pearson correlation between regional traffic data of total timespan, $d_{i,j}$ is an Euclidean distance, and $\sigma_\text{dist}$ is a standard deviation of distances.

\subsubsection{Spatial Attention}
Basically, a spatial attention and a temporal attention are in the same form, but in different attentional dimension: whether in spatial or temporal.
A spatial attention is used for road-level spatial feature extraction by calculating $\mathbb{R} ^ {N_X \times N_X}$ attention score to find correlation between the roads.
%
% Note that there is no spatial attention in region-level computation.
%
% This is because, it is barely impossible to calculate attention scores for all the pairs of lte cells.
We first concat $\text{STE}_X$ to the previous hidden output, and apply a self attention as $ H_{XS}^{(l)} = \text{SelfAtt} (( H_{X}^{(l-1)} \ \Vert \  \text{STE}_X )_S)$.

\subsubsection{Temporal Attention}

The temporal attention is used for both region-level and road-level temporal feature extraction.
%
% This module is same as temporal attention, except that it conducts attention on spatial dimension.
%
The module calculates $\mathbb{R} ^ {P \times P}$ or $\mathbb{R} ^ {Q \times Q}$ attention scores depending on 
%whether the module is used in 
the module's objective is an encoder or decoder. %, respectively.
%
% Note that there is no spatial attention in region-level computation.
%
% This is because, it is barely impossible to calculate attention scores for all the pairs of lte cells.
Similar to spatial attention, we first concatenate $\text{STE}$ to the input before we apply temporal attention.
For each modality, we concatenate corresponding $\text{STE}$ with previous hidden output and apply a self attention as $ H_{MT}^{(l)} = \text{SelfAtt}(( H_{M}^{(l-1)} \ \Vert \  \text{STE}_M )_T)$, where $M \in \{\text{`X'}, \text{`Z'} \}$.

\subsubsection{Gated Fusion}
In order to mix spatial and temporal hidden outputs depending on their importance, we leverage gated fusion proposed by \citeauthor{zheng2020gman}. %\todo{We leave this definition to the original GMAN paper due to  lack of pages.}

\begin{equation}
\label{eq:gatedfusion}
\begin{gathered}
    H^{(l)} =  P \ast H_S^{(l-1)} + (1-P) \ast H_T^{(l-1)}, \ \  \text{with}\\
% \end{equation}
% \begin{equation}
    P = \text{ReLU}(H_S^{(l-1)} \mathbf{W}_{p,1} + H_T^{(l-1)} \mathbf{W}_{p,2} + \mathbf{b}_p)
\end{gathered}
\end{equation}

where $\ast$ is an element-wise multiplication and $\mathbf{W}_{p,1} \in \mathbb{R}^{D \times D}$, $\mathbf{W}_{p,2} \in \mathbb{R}^{D \times D}$, $\mathbf{b}_{p} \in \mathbb{R}^{D}$ are trainable parameters. A gated fusion is used both for R-STAtt and G-STAtt blocks.

% \begin{equation}
% \label{eq:gatedfusion}
% \begin{gathered}
%     H^{(l)} =  p \star H_S^{(l)} + (1-p) \star H_T^{(l)}, where \\
%     z = \text{ReLU}(H_S^{(l)} W_{p,1} + H_T^{(l)} W_{p,2} + b_p)
% \end{gathered}
% \end{equation}

\subsection{Bipartite Spatial Transform Attention}

% \begin{figure}[h]
%     \centering
%     \includegraphics[width=8cm]{bipartite.png}
%     \caption{Bipartite spatial transform attention}
%     \label{fig:bipartite}
% \end{figure}

To give real-time regional knowledge for road-level traffic prediction, we propose a bipartite transform attention that converts the spatial modality of regional representations  %resulted from R-encoder and R-decoder 
from $N_Z$ to $N_X$. Since a road is affected by its nearby built-in environment~\cite{cervero1997travel}, we incorporate proximity information on a masked attention network.
%도로가 지나가는 region cell과 도로간의 bipartite 네트워크를 구성하여, region-level spatio dimension에서 road-level spatio dimension으로 transform 하는 attention 네트워크이다. 모델에서는 ZP를 STERP를 통해 STEGP로 바꾸거나 ZQ를 STERQ를 통해 STEGQ로 바꾼다.
%
%For bipartite spatial transformation, we leverage masked attention network as explained in Equation \ref{eq:maskattn}. 
We give soft gaussian mask $M \in \mathbb{R}^{N_X \times N_Z}$ intuitioned by \citeauthor{sperber2018self} as follows: % $M_{r,c} = - {d_{r,c}^2}/{2\sigma_r^2}$ using distance between $r$-th road and $c$-th regional cell.

\begin{equation}
\label{eqn:mask}
    M_{r,c} = - d_{r,c}^2 / 2 \sigma_r^2
\end{equation}

% \begin{equation}
% \label{eqn:mask}
%     M_{r,c} = - \frac{d_{r,c}^2}{2\sigma_r^2}
% \end{equation}
%
%where $d_{r,c}$ is distance between r-th road and c-th region cell, 
%
Here, $\sigma_r \in \mathbb{R}$ is a trainable standard deviation of distances to regional cells from $r$-th road. This enables individual road to train how much proximity information to take. Furthermore, we extend this approach into multi-head masked attention by applying different attention masks for $K$-heads that trains $\boldsymbol{\sigma} \in \mathbb{R}^{N_X \times K}$. We formulate this 
%bipartite spatial transform attention that outputs 
road-level real-time regional knowledge representation $HK_{T}$ as follows:

% \begin{equation}
\begin{multline}\label{eq:bitrans}
    {HK}_{T} = \text{MHAtt}^M \bigl( f_1(\text{STE}_{ZT}^{(l-1)}), f_2(H_{ZT}^{(l-1)}), \\  
    f_3(H_{Z}^{(l-1)}), M_{1,...,K} \bigr), \ \ \text{where} \ \  T \in \{P, Q\}.
\end{multline}
% \end{equation}

\subsection{Temporal Transform Attention}
In order to transform the temporal sequence from $P$ to $Q$ for an actual next sequence prediction, we leverage temporal transform attention. As described in Figure~\ref{fig:method}, we use $P$ and $Q$ temporal sequences of $\text{STE}$ or $HK$ as query and key to calculate $\mathbb{R} ^ {Q \times P}$ attention scores for each modality, and use previous hidden output as value, as follows: % defined in Equation \ref{eq:ttrans}. % Note that each attention score is calculated as $\mathbb{R} ^ {Q \times P}$ in both cases.

\begin{equation}\label{eq:ttrans}
\begin{gathered}
    H_{Z}^{(l)} = \text{MHAtt}(f_1(\text{STE}_{ZQ}^{(l-1)}), f_2(\text{STE}_{ZP}^{(l-1)}), f_3(H_{ZP}^{(l-1)})) \\ 
    H_{X}^{(l)} = \text{MHAtt}(f_1({HK}_{Q}^{(l-1)}), f_2({HK}_{P}^{(l-1)}), f_3(H_{XP}^{(l-1)}))
\end{gathered}
\end{equation}

\subsection{Objective function}

We insert 2-layer FCNs at the beginning when we expand dimension from 1 to $D$, and at the end when we shrink to produce the final output from $D$ to 1 as described in Figure~\ref{fig:method}. Using this final output prediction $\hat{Y}$, we train our model using mean absolute error with the ground truth $Y \in \mathbb{R}^{Q\times N_X}$ for the objective loss function defined as  % defined as Equation \ref{eq:objfunc}. 
%\begin{equation}
%\label{eq:objfunc}
$    \mathcal{L}(\theta) = \frac{1}{N_X Q } {\sum_{t=t_{P+1}}^{t_{P+Q}} | Y_t - \hat{Y}_t|}$.
%\end{equation}

\section{Dataset}

Seoul provides road traffic speed\footnote{https://topis.seoul.go.kr/refRoom/openRefRoom\_1.do} and regional living population data\footnote{http://data.seoul.go.kr/dataList/OA-14979/F/1/datasetView.do} for public data use policy as Figure~\ref{fig:overview}.
We construct our real-world urban traffic dataset in three representative downtown regions in Seoul, Korea: Gangnam, Hongik, and Jamsil. %\js{why choosing these three regions?} %, where commercial and residential areas are mixed.
The dataset description is listed in Table~\ref{tab:dataset}.

\subsubsection{Road traffic speed data}

Road traffic speed is collected by measuring average hourly traffic speed from GPS signals of around 70,000 taxis in Seoul.
We construct a road network using geographic information system (GIS).
Each node represents a road while each edge does a link between given two roads.
We assume that a traffic sensor is located at the center of a road, and calculate a distance to an adjacent road. % using road lengths.
%We find well-known benchmark dataset such as METR-LA or PemS-Bay uses edge information although they are not directly adjacent.
%For fairness to construct our dataset with other baseline models that are highly dependent on road connection, we also add edges that are reachable within 1.5 km by extra walks.

\subsubsection{Real-time regional living population (LTE)}
We use regional living population estimated hourly from an LTE cell tower. %, and officially published by Seoul Metropolitan Government.
The original data is estimated in zip-code based region unit, but we spatially normalize into 150-meter grid cells. 
%
%Although this dataset does not show the exact number of people in the region due to vendor or user specific,
%Since the LTE data can be depend on the vendor and does not exactly calculate the number of people at the timestamp,
We normalize cell data using training data by the Z-score method, and use this trend data for our model.

\begin{table}[t]
\centering
\begin{tabular}{c|c|c|c}
\hline
Region & Gangnam & Hongik & Jamsil \\ \hline
\#Nodes ($N_X$) & 148 & 93 & 134 \\
\#Edges & 1334 & 937 & 1117 \\ 
Avg. Dist. & 612 & 505 & 530 \\ \hline
$N_Z = N_h \times N_w$ & 34 × 33 & 22 × 27 & 23 × 39 \\  
Avg. \#POIs & 79.8 &  55.6 &  32.1 \\ \hline
Timespan & \multicolumn{3}{c}{3/1/2018 $\sim$8/31/2018 (1h, 4392 steps)} \\ \hline 
\end{tabular}
\caption{Dataset description. Each row means the number of nodes, edges, and average distance between the nodes, the number of regional cells, the average number of POIs per cell, and total timespan.}
\label{tab:dataset}
\end{table}

\subsubsection{Point of Interests (POIs)}

We count the number of POIs using the business registration database\footnote{https://www.localdata.go.kr/devcenter/dataDown.do}.
The original data consists of more than 300 subcategories, but we reduce it to 10 representative categories: \textit{shopping, food, cafe, beauty, work, hospital, school, art\&entertainment, lodging}, and \textit{nightlife} 
based on Foursquare taxonomy\footnote{https://developer.foursquare.com/docs/build-with-foursquare/categories/}. 
Table~\ref{tab:roadpoi} briefly shows the built-in environment of roads in each region. 
Gangnam is the most crowded region where every type of POI, including commercial and workplaces, is located. 
Hongik is less crowded, but still many shops and restaurants are located as well as lodging places for travelers. 
Jamsil has the least POIs, while a few entertainments such as a stadium and an amusement park are located, although it is not revealed in the table. Many parts of Jamsil consist of residential areas as the high number of schools represents this.% characteristic. 

\begin{table}[h]
\begin{tabular}{|r|r|r|r|r|r|r|r|}
\hline
        & shop. & work  & food  & a\&e  & lodg. & night. & sch. \\ \hline
G & \textbf{41.7} & \textbf{24.5} & 15.1 & \textbf{2.44} & 0.31  & \textbf{0.78}   & 0.39 \\ \hline
H  & 27.8 & 21.5 & \textbf{20.2} & 1.89 & \textbf{1.56}  & 0.39   & 0.28 \\ \hline
J  & 17.3 & 7.1  & 8.2  & 1.91 & 0.33  & 0.48   & \textbf{0.45} \\ \hline
\end{tabular}
\caption{Mean POI distribution of roads, where POI density for each road is calculated by averaging the POI distribution of 500m nearby cells. (Here, three other POIs are skipped.) }\label{tab:roadpoi}
\end{table}

\begin{table*}[t]
\begin{tabular}{|c|l||c|c|c||c|c|c|}
\hline
                         & Model        & 1h MAE        & 2h MAE        & 3h MAE        & Avg. MAE      & Avg. RMSE     & Avg. MAPE       \\ \hline\hline
\multirow{7}{*}{\rotatebox[origin=c]{90}{Gangnam}} & HA      & 1.500         & 1.500         & 1.499         & 1.500         & 2.138         & 8.45\%          \\ \cline{2-8} 
                         & SVR/+LTE     & 1.505 / 1.436 & 1.835 / 1.688 & 1.959 / 1.780 & 1.766 / 1.635 & 2.467 / 2.279 & 9.91\% / 9.14\% \\ \cline{2-8} 
                         & RFR/+LTE     & 1.328 / 1.406 & 1.522 / 1.588 & 1.593 / 1.663 & 1.481 / 1.552 & 2.045 / 2.148 & 8.20\% / 8.55\% \\ \cline{2-8} 
                         & DCRNN/+LTE   & 1.299 / 1.339 & 1.516 / 1.558 & 1.577 / 1.618 & 1.464 / 1.505 & 1.998 / 2.051 & 7.97\% / 8.18\% \\ \cline{2-8} 
                         & ST-Meta/+LTE & 1.284 / 1.278 & 1.486 / 1.479 & 1.541 / 1.542 & 1.437 / 1.433 & 1.968 / 1.964 & 7.83\% / 7.83\% \\ \cline{2-8} 
                         & GMAN/+LTE    & 1.319 / 1.314 & 1.407 / 1.402 & 1.444 / 1.438 & 1.390 / 1.385 & 1.904 / 1.897 & 7.56\% / 7.57\% \\ \cline{2-8} 
                         & OURS         & \textbf{1.265}($\pm$.006)         & \textbf{1.387}($\pm$.005)         & \textbf{1.433}($\pm$.004)         & \textbf{1.362}($\pm$.005)          & \textbf{1.863}($\pm$.005)         & \textbf{7.38\%}($\pm$.01\%)          \\ \hline\hline

                        %  & OURS         & \textbf{1.266}         & \textbf{1.381}         & \textbf{1.427}         & \textbf{1.358}         & \textbf{1.864}         & \textbf{7.40\%}          \\ \hline\hline
\multirow{7}{*}{\rotatebox[origin=c]{90}{Hongik}}  & HA     & 1.515         & 1.515         & \textbf{1.515}         & 1.515         & 2.375         & 7.84\%          \\ \cline{2-8} 
                         & SVR/+LTE     & 1.542 / 1.534 & 1.796 / 1.738 & 1.887 / 1.810 & 1.742 / 1.694 & 2.721 / 2.654 & 9.06\% / 8.80\% \\ \cline{2-8} 
                         & RFR/+LTE     & 1.410 / 1.494 & 1.573 / 1.648 & 1.630 / 1.699 & 1.538 / 1.614 & 2.262 / 2.369 & 8.01\% / 8.45\% \\ \cline{2-8} 
                         & DCRNN/+LTE   & \textbf{1.373} / 1.379 & 1.516 / 1.531 & 1.559 / 1.573 & 1.482 / 1.494 & 2.227 / 2.262 & 7.67\% / 7.72\% \\ \cline{2-8} 
                         & ST-Meta/+LTE & 1.386 / 1.380 & 1.543 / 1.535 & 1.593 / 1.579 & 1.507 / 1.498 & 2.282 / 2.258 & 7.77\% / 7.74\% \\ \cline{2-8} 
                         & GMAN/+LTE    & 1.436 / 1.429 & 1.504 / 1.493 & 1.531 / 1.523 & 1.491 / 1.481 & 2.239 / 2.212 & 7.72\% / 7.68\% \\ \cline{2-8} 
                         & OURS         & 1.399($\pm$.004)         & \textbf{1.490}($\pm$.006)         & 1.524($\pm$.007)         & \textbf{1.471}($\pm$.005)         & \textbf{2.200}($\pm$.011)         & \textbf{7.60\%}($\pm$.06\%)          \\ \hline\hline
                        %  & OURS         & 1.396         & \textbf{1.484}         & 1.516         & \textbf{1.465}         & \textbf{2.194}         & \textbf{7.55\%}          \\ \hline\hline
\multirow{7}{*}{\rotatebox[origin=c]{90}{Jamsil}}  & HA       & 1.734         & 1.734         & 1.734         & 1.734         & 2.409         & 7.83\%          \\ \cline{2-8} 
                         & SVR/+LTE     & 1.657 / 1.675 & 1.941 / 1.907 & 2.060 / 2.014 & 1.886 / 1.866 & 2.658 / 2.599 & 8.34\% / 8.30\% \\ \cline{2-8} 
                         & RFR/+LTE     & 1.536 / 1.612 & 1.728 / 1.820 & 1.806 / 1.913 & 1.690 / 1.782 & 2.360 / 2.489 & 7.46\% / 7.85\% \\ \cline{2-8} 
                         & DCRNN/+LTE   & 1.487 / 1.481 & 1.663 / 1.656 & 1.720 / 1.713 & 1.623 / \textbf{1.617} & 2.264 / 2.258 & 7.19\% / 7.16\% \\ \cline{2-8} 
                         & ST-Meta/+LTE & 1.481 / \textbf{1.478} & 1.662 / 1.662 & 1.731 / 1.726 & 1.625 / 1.622 & 2.268 / 2.259 & 7.16\% / 7.16\% \\ \cline{2-8} 
                         & GMAN/+LTE    & 1.574 / 1.580 & 1.674 / 1.673 & 1.726 / 1.720 & 1.658 / 1.658 & 2.306 / 2.302 & 7.31\% / 7.31\% \\ \cline{2-8} 
                         & OURS         & 1.523($\pm$.006)         & \textbf{1.647}($\pm$.009)         & \textbf{1.702}($\pm$.008)         & 1.624($\pm$.007)         & \textbf{2.258}($\pm$.010)         & \textbf{7.13\%}($\pm$.03\%)          \\ \hline
\end{tabular}\caption{Performance comparison of different road-level traffic prediction on our datasets (lower is better).}\label{tab:result}
\end{table*}

\subsubsection{Satellite image features}
We extract satellite image features using QGIS VWorld satellite plugin\footnote{https://dev.vworld.kr/dev/v4api.do}. 
We combine satellite image patches of cells in three regions, and train a convolutional auto-encoder using 90\% of training data and 10\% of validation data. Each image patch is resized in $36\times36$ size, and we use two stacks of $3\times3$-kernel CNN and CNNtranspose for encoder and decoder. Then we extract the encoded features using its encoder part and reduce the dimension into $D$ by principal component analysis.

% \begin{figure}[h]
%     \centering
%     \includegraphics[width=8cm]{similar_patches.png}
%     \caption{Similar patches clustered by extracting latent features from pretrained autoencoder, and conducting PCA and K-means clustering. }
%     \label{fig:stablock}
% \end{figure}
% train auto-encoder and extract latent feature. Then conduct PCA to reduce dimension into D. 

\section{Experiment}

\subsection{Experimental Settings}

\subsubsection{Data Processing}
We split the dataset successively in 70\% of the data for training, 10\% of the data for validation, and 20\% of the data for testing.
%For dynamic convolution, we calculate pearson correlation between region cells using only training dataset.
We find that the missing values in training data affect the results of all baseline models poorly, so we fill them by next or previous valid values.
%However, we do not measure metrics on missing values in testing data for evaluation.
%본 실험에서 실험 결과가 missing value에 크게 영향을 받는것을 발견하였다. 그리고 training data의 missing value를 인근 데이터로 대체하여 학습한 결과 전 모델의 성능이 향상되었다. 본 실험은 missing value 처리가 된 상태로 진행하였으며, evaluation은 missing value는 빼고 metric을 계산하였다.

\subsubsection{Hyperparamters}

The model inputs $P = 12$ historical time steps (12 hours) to predict the next $Q=3$ steps (3 hours) of road traffic values. The hyperparameters for our model are set as $L_X = 3$, $L_Z=2$, $K=8$, $d_h=8$ $(D=64)$, and $\lambda_r=0.6$. The initial learning rate is set to 0.001, and we use Adam optimizer~\cite{kingma2014adam} to train our model. Our model is trained on one GPU environment of NVIDIA GeForce GTX 1080 Ti. We test 5 times and record the average and standard deviation of results.

\subsubsection{Metrics}
We use three major traffic evaluation metrics: Mean Absolute Error (MAE), Root Mean Squared Error (RMSE), and Mean Absolute Percentage Error (MAPE).

\subsubsection{Baselines}

We set our baselines as 
(1) Historical Average based on the hour-of-day and day-of-week (\textbf{HA}), 
(2) Support Vector Regression (\textbf{SVR}), 
(3) Random Forest Regression (\textbf{RFR}), 
(4) \textbf{DCRNN}~\cite{li2017diffusion}, %\footnote{https://github.com/liyaguang/DCRNN} 
(5) \textbf{ST-MetaNet}~\cite{pan2019urban_stmetanet}, %\footnote{https://github.com/panzheyi/ST-MetaNet} 
(6) \textbf{GMAN}~\cite{zheng2020gman}. %\footnote{https://github.com/zhengchuanpan/GMAN} 
%
%We use sklearn library\footnote{https://scikit-learn.org/} for SVR and RFR, and we use default settings for each baseline model.
Unlike GMAN, other models does not have an extra module to embed a temporal timestamp.
For SVR and RFR, we give the hour-of-day and the day-of-week of the first timestamp of P input sequences as extra features.
For DCRNN, ST-MetaNet, we give these two timestamp information as an additional input, total 3 input channels.
For each baseline models, we also test with road-level regional population data by averaging cells within 500 meters (walking distance) from each road, and give as an extra input channel.
In this case, the input channel of road traffic data becomes 4 for DCRNN and ST-MetaNet, and 2 for GMAN.

\begin{table}[h]
\begin{tabular}{|c|c|c|c|c|}
\hline
 & DCRNN & ST-Meta & GMAN & Ours \\ \hline
\# Params & 373,312 & 83,717 & 883,905 & 842,177 \\ \hline
Train (s/ep.) & 34.0 & 44.9 & 94.1 & 236.8 \\ \hline
\# Epochs & 188 & 206 & 36 & 61 \\ \hline
\end{tabular} \label{tab:check}\caption{The number of trainable parameters, training time, and total epochs of each model on Gangnam dataset.}
\end{table}

\subsection{Experimental results}

\subsubsection{Forecasting Performance Comparison}
%\textcolor{red}{[This section needs to be combined with the third one, which will justify why we use place sociality, that is, how regions being used and their correlations with road traffic condition along with certain seasonal, calendar, or social data]}

Table~\ref{tab:result} shows experimental results on our dataset. We denote +LTE as the model with extra LTE input. 
%Since we calculate historical average based on the weekly and hourly trend, the error of HA shows the prediction difficulty in each region.
%
First of all, our model outperforms the other baselines on average RMSE, and MAPE.
Compared to the RNN-based models (DCRNN and ST-MetaNet), the attention based models (GMAN and our model) are better at longer step prediction (2h, 3h).
%
%However, even at 1h-step prediction, our model shows best at Gangnam and marginally good as baselines in other regions.
%
Considering that GMAN is also an attention-based model, the result shows that the improvement of our model is driven by the real-time regional knowledge.
In ST-MetaNet and GMAN, the LTE data gives slight more information to improve prediction. However, in DCRNN, it leads to worse prediction in Gangnam and Hongik.
Since DCRNN depends more on graph computation, it cannot fully utilize the regional LTE data due to different original modality.
On the other hand, ST-MetaNet trains individual parameters for an RNN cell of each traffic sensor and GMAN applies spatial attention, but without an strict attention mask for edge connection, thus they can slightly benefit from LTE data.

\begin{figure}[t]
\centering
\subfloat[Gangnam (POI-H)]{\includegraphics[width=0.23\textwidth]{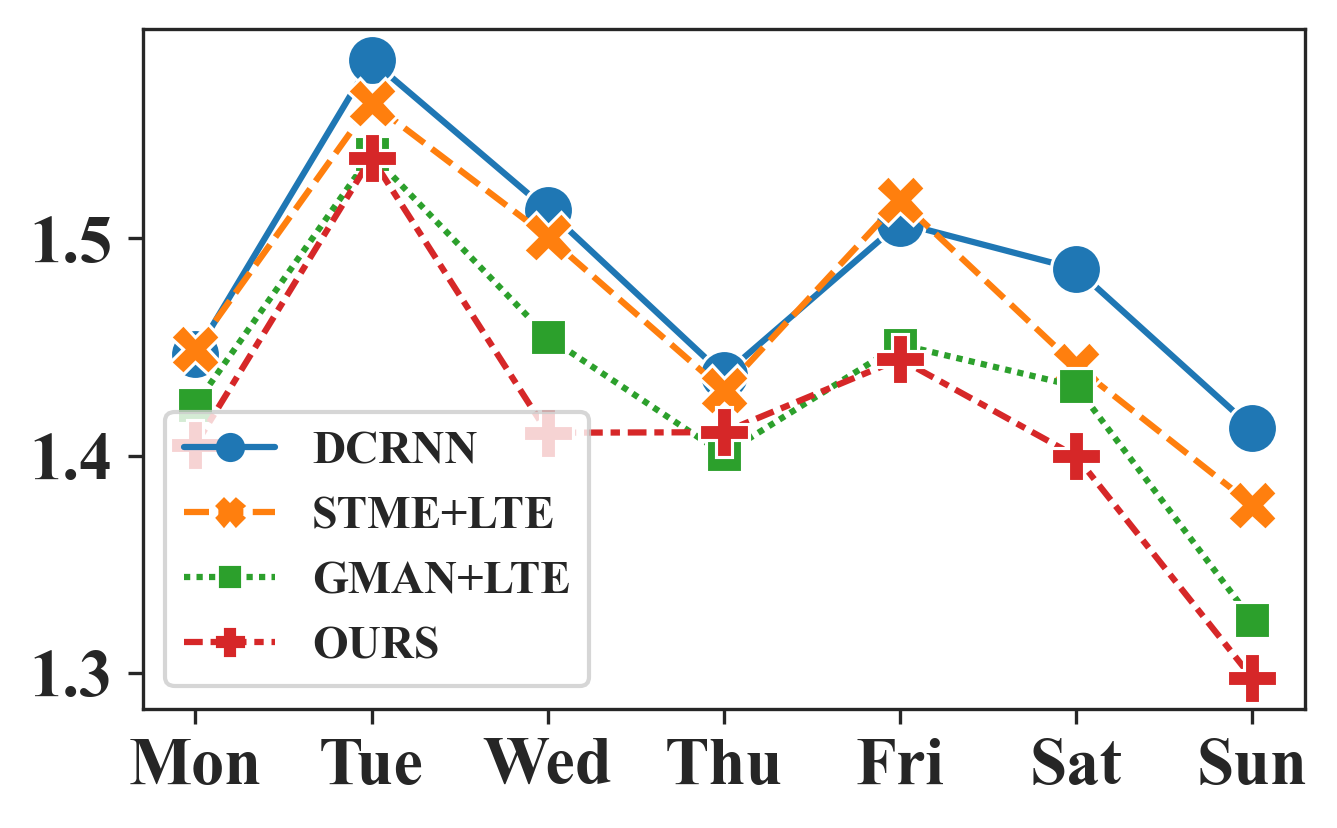}} \null\hfill
\subfloat[Gangnam (POI-L)]{\includegraphics[width=0.23\textwidth]{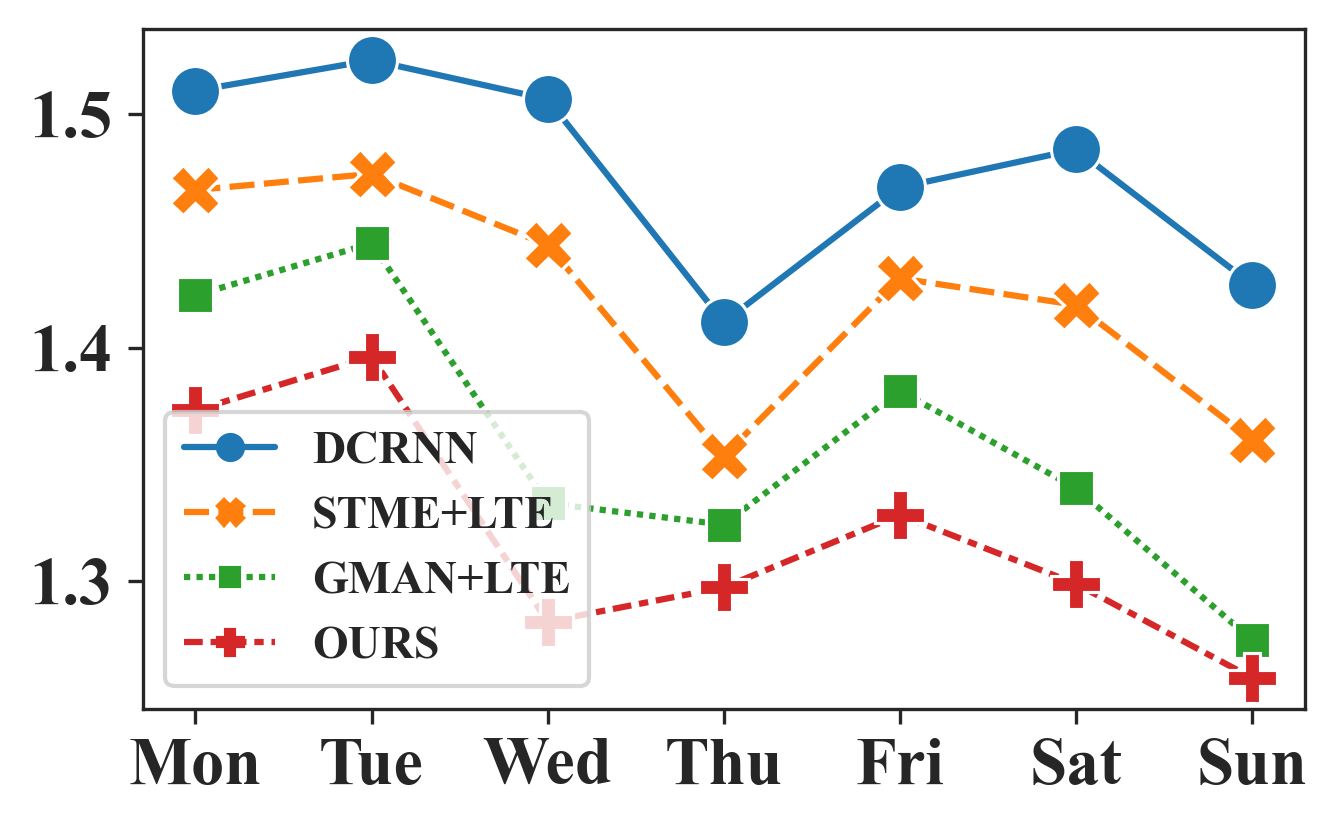}} \hfill
\subfloat[Hongik (POI-H)]{\includegraphics[width=0.23\textwidth]{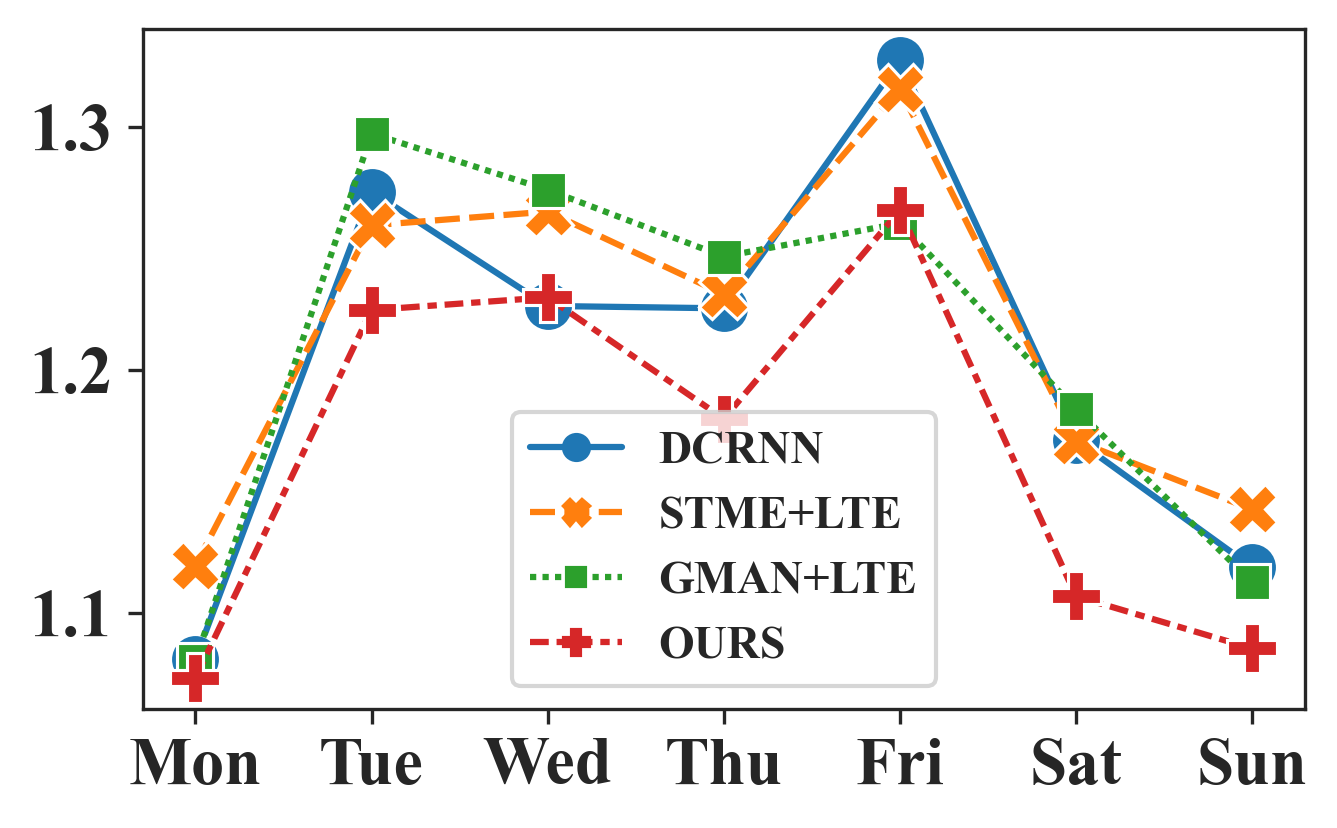}} \null\hfill
\subfloat[Hongik (POI-L)]{\includegraphics[width=0.23\textwidth]{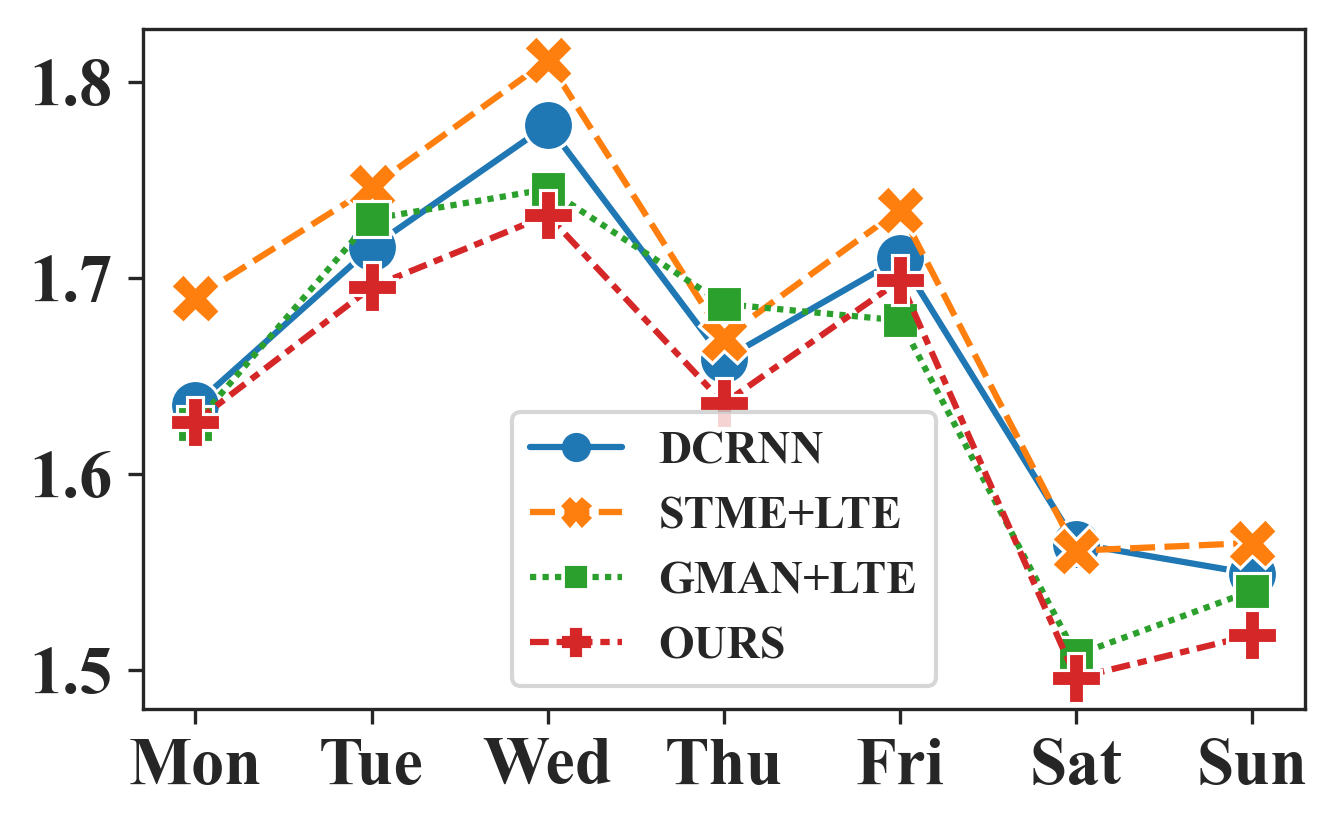}} \hfill
\subfloat[Jamsil (POI-H)]{\includegraphics[width=0.23\textwidth]{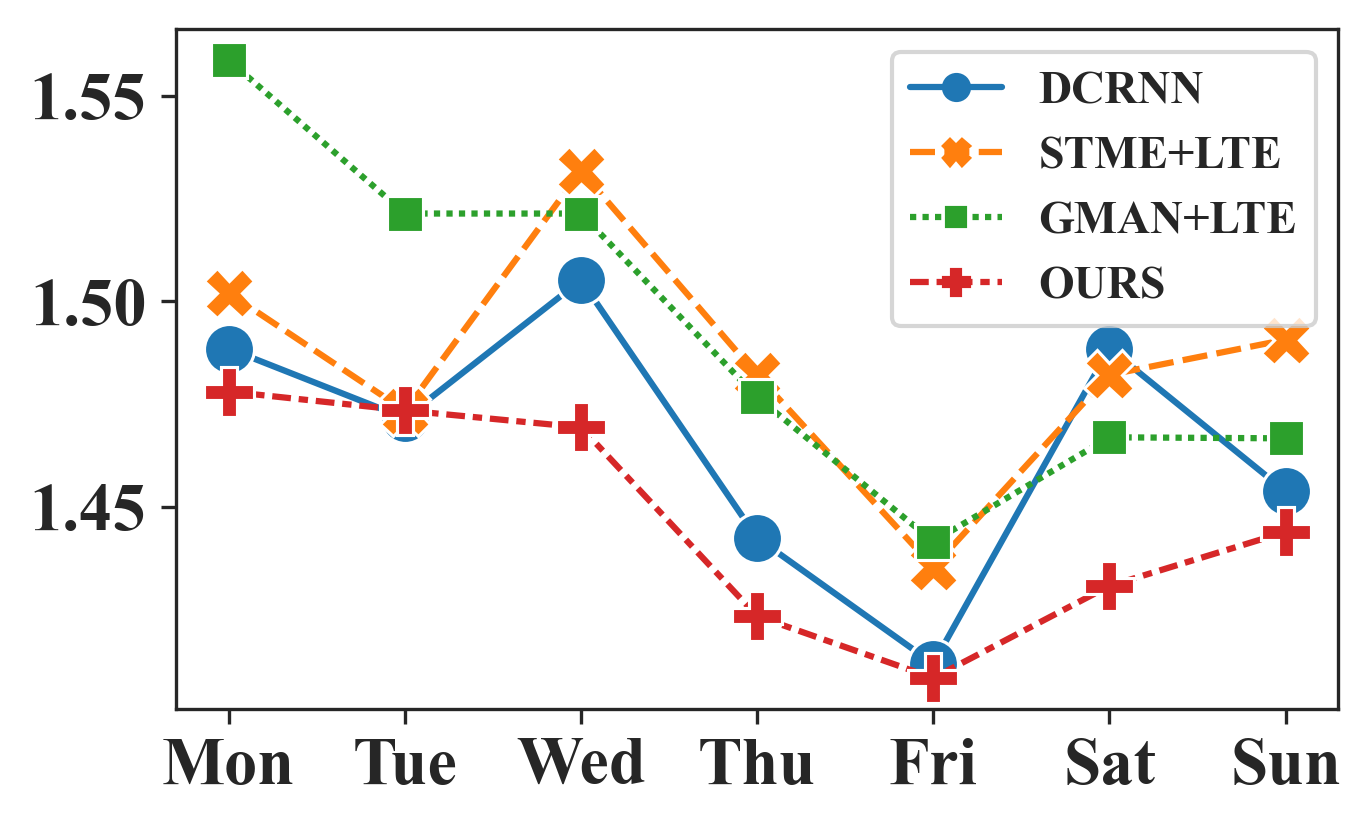}} \null\hfill
\subfloat[Jamsil (POI-L)]{\includegraphics[width=0.23\textwidth]{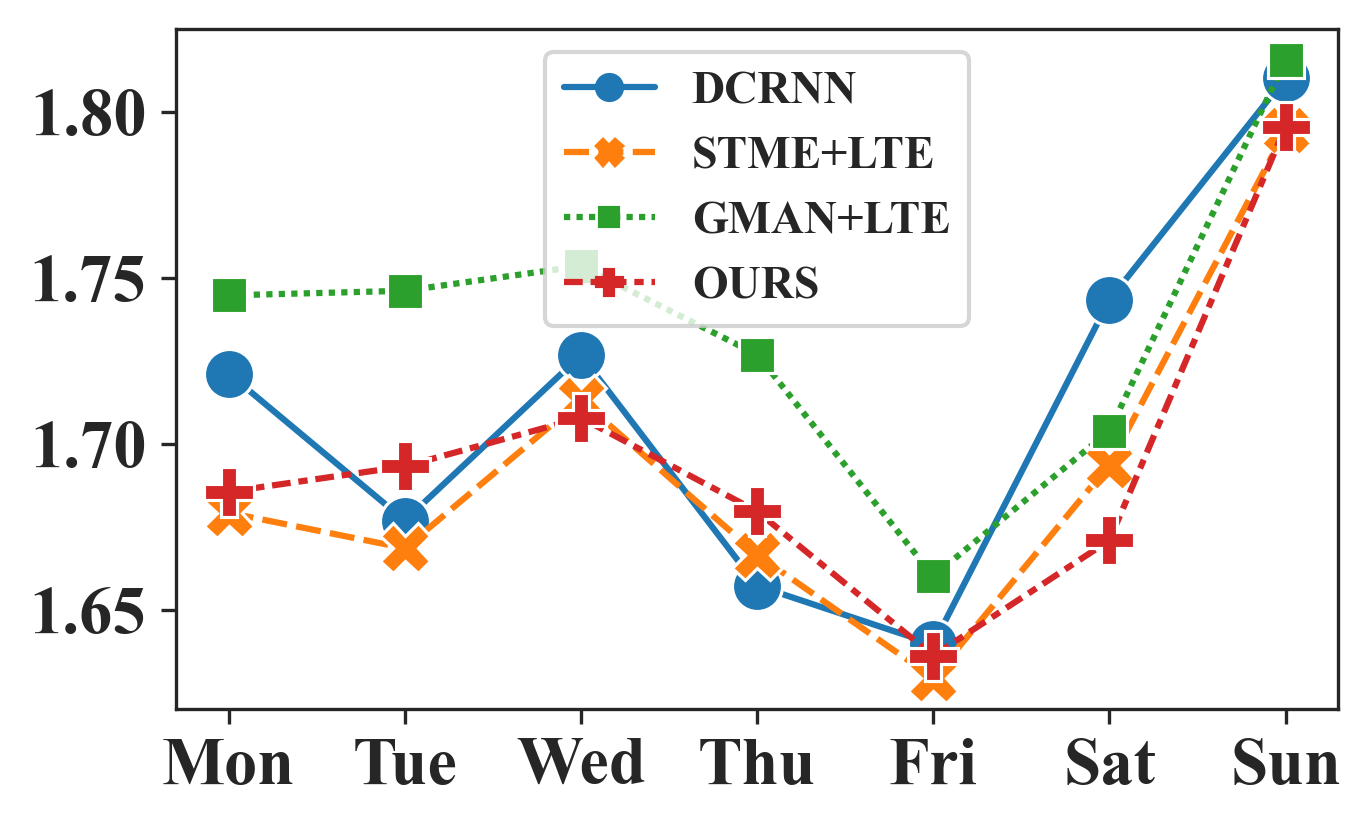}} \hfill
    \caption{Weekly analysis of performance on different roads of POI densities (average MAE, model improvement compared to GMAN+LTE).}\label{fig:weekly}
\end{figure}

% \begin{figure}[t]
%     \centering
%     \includegraphics[width=8.4cm]{figures/roadpoi.png}
%     \caption{Weekly analysis of performance on different roads classified by POI densities (average MAE). Negative percentage numbers represent model improvement.}
%     \label{fig:weekly}
% \end{figure}

% \begin{figure}[h]
%     \centering
%     \includegraphics[width=8cm]{figures/geographic.png}
%     \caption{Examples of road congestion prediction on each region (1h-step MAE).}
%     \label{fig:geographic}
% \end{figure}
% % 66, 2, 14

\subsubsection{Weekly Analysis on Different Road POI Characteristics}

To show how regional knowledge benefits our model, we compare two groups of roads from each region---top 30\% of high POI density (\textbf{POI-H}) and bottom 30\% of low POI density (\textbf{POI-L})---and conduct a weekly comparison in Figure~\ref{fig:weekly}.
We compare our model with DCRNN, ST-MetaNet+LTE, GMAN+LTE, considering whether they can benefit from LTE data.
In general, our model outperforms more in POI-H than in POI-L compare to other baselines.
Here, we also measure the weekly maximal performance improvement compare to GMAN+LTE to show how much our model benefits from regional knowledge.
In Gangnam, our model is improved up to 3.0\% in POI-H and 3.8\% in POI-L (Figure~\ref{fig:weekly}~(a,b)).
When we review Table~\ref{tab:roadpoi}, the POI density is highest in Gangnam among the test regions, so there are still many POIs in POI-L group. This allows our model to perform well on all roads in Gangnam.
In Hongik, our model is improved up to 6.5\% in POI-H and 3.0\% in POI-L (Figure~\ref{fig:weekly}~(c,d)). %
Hongik is visited by people who come for dining and socializing and travelers who come for entertainment. Thus our model primarily benefits on the roads nearby POIs.
% \todo{Since there are more social events in Jamsil, especially near the stadium and amusement park \footnote{https://stadium.seoul.go.kr/board/show-event}, this dynamic and social regional knowledge helps greatly in our model.}
% %
In Jamsil, our model is improved up to 5.2\% in POI-H and 3.4\% in POI-L (Figure~\ref{fig:weekly}~(e,f)).
Since there is a big stadium\footnote{https://stadium.seoul.go.kr/board/show-event} and an amusement park\footnote{https://adventure.lotteworld.com/eng} where dynamic social events occur, our model can capture such live regional information to improve our model.
On the other hand, the low-POI roads in Hongik and Jamsil are located in residences or freeways, so there is less dynamic regional knowledge that our model can have benefit.

%Except for Gangnam, the result shows that our model predicts more accurately on high POI roads in Hongik. 
%
%since thus there is useful regional information on these roads that can help our model.
%
% In Hongik, our model performs better in high-POI roads, since this region is 
% https://stadium.seoul.go.kr/board/show-event
%
%Also, since many restaurants are closing at Monday as their weekly holiday and re-open from Tuesday, our model seems better starting from Tuesday.
%

\subsubsection{Ablation Study}

% \begin{table}[h]
% \centering
% \begin{tabular}{|c|c|c|c|c|c||c|c|c|}
% \hline
% \# & L. & P. & S. & D. & M. & Gang. & Hong. & Jam. \\ \hline \hline
% a & &  &  & 0.7 &  & \textbf{1.353} & 1.471 & 1.623 \\ \hline
% b & &  &  & 0.5 &  & 1.369 & 1.472 & 1.649 \\ \hline
% c & &  &  & 0.6 &  & 1.358 & \textbf{1.465} & 1.615 \\ \hline
% d & & X & X & 0.6 &  & 1.512 & 1.596 & 1.734 \\ \hline
% e &  & X &  & 0.6 &  & 1.363 & 1.472 & \textbf{1.605} \\ \hline
% f & &  & X & 0.6 &  & 1.359 & 1.472 & 1.627 \\ \hline
% g & X &  &  & X &  & 1.383 & 1.500 & 1.642 \\ \hline
% h & &  &  & X &  & 1.372 & 1.482 & 1.621 \\ \hline
% i & &  &  & 0.6 & X & 1.372 & 1.492 & 1.646 \\ \hline
% % &  &  & X & X & 1.377 & 1.487 & 1.644 \\ \hline
% \end{tabular}
% \caption{Effectiveness of each module measured in average MAE. Each column stands for: LTE (\textbf{L}), POI (\textbf{P}), Satellite (\textbf{SAT}), Dynamic convolution and $\lambda_r$-value (\textbf{D}), and the making of a bipartite attention (\textbf{M}).}\label{tab:ablation}
% \end{table}

\begin{table}[h]
\begin{tabular}{|c|c|c|c|c|c||c|c|c|}
\hline
\# & L. & P. & S. & D.  & M. & Gang. & Hong. & Jam.  \\ \hline\hline
a  &    &    &    & 0.7 &    & 1.365 & 1.476 & 1.629 \\ \hline
b  &    &    &    & 0.5 &    & 1.366 & 1.472 & 1.635 \\ \hline
c  &    &    &    & 0.6 &    & 1.361 & 1.470 & 1.623 \\ \hline
d  &    & X  & X  & 0.6 &    & 1.549 & 1.582 & 1.729 \\ \hline
e  &    & X  &    & 0.6 &    & 1.363 & 1.471 & 1.624 \\ \hline
f  &    &    & X  & 0.6 &    & 1.364 & 1.470 & 1.628 \\ \hline
g  & X  &    &    & X   &    & 1.417 & 1.500 & 1.647 \\ \hline
h  &    &    &    & X   &    & 1.367 & 1.474 & 1.629 \\ \hline
i  &    &    &    & 0.6 & X  & 1.372 & 1.486 & 1.629 \\ \hline
\end{tabular}
\caption{Effectiveness of each module measured in average MAE. Each column stands for: LTE (\textbf{L}), POI (\textbf{P}), Satellite (\textbf{SAT}), Dynamic convolution and $\lambda_r$-value (\textbf{D}), and the making of a bipartite attention (\textbf{M}).}\label{tab:ablation}
\end{table}

% \begin{table}[h]
% \begin{tabular}{|c|c|c|c|c|c||c|c|c|}
% \hline
% \# & L. & P. & S. & D.  & M. & Gang. & Hong. & Jam.  \\ \hline\hline
% a  &    &    &    & 0.7 &    & 1.353 & 1.470 & 1.618 \\ \hline
% b  &    &    &    & 0.5 &    & 1.356 & 1.466 & 1.614 \\ \hline
% c  &    &    &    & 0.6 &    & 1.354 & 1.462 & 1.614 \\ \hline
% d  &    & X  & X  & 0.6 &    & 1.437 & 1.547 & 1.706 \\ \hline
% e  &    & X  &    & 0.6 &    & 1.354 & 1.460 & 1.605 \\ \hline
% f  &    &    & X  & 0.6 &    & 1.359 & 1.465 & 1.616 \\ \hline
% g  & X  &    &    & 0.6 &    & 1.374 & 1.494 & 1.631 \\ \hline
% h  &    &    &    & X   &    & 1.361 & 1.469 & 1.621 \\ \hline
% i  &    &    &    & 0.6 & X  & 1.359 & 1.482 & 1.614 \\ \hline
% \end{tabular}
% \caption{Effectiveness of each module measured in average MAE. Each column stands for: LTE (\textbf{L}), POI (\textbf{P}), Satellite (\textbf{SAT}), Dynamic convolution and $\lambda_r$-value (\textbf{D}), and the making of a bipartite attention (\textbf{M}).}\label{tab:ablation}
% \end{table}

Table~\ref{tab:ablation} shows effectiveness of each modules of our method. 
%Each column represents LTE, POI, Satellite image, Dynamic convolution, and making of Bipartite attention.
%
We first empirically measure $\lambda_r$-value is suitable around 0.6 (\#a, \#b, \#c). % and conduct ablation study on this basic setting.
When we train our model only with geographic location without POI or satellite image information, the performance gets much worse as LTE data misleads the prediction (\#d). 
However, if we have either POI or satellite image, it gives the POI or land use information that is useful to analyze regional LTE pattern (\#e, \#f). 
When we do not consider LTE data, we simply conduct $L_z$ stacks of $5 \times 5$-kernel 2D-CNN using geographic regional features, and produce static regional knowledge to feed bipartite spatial attention (\#g). 
% without , %as 2D grid image of region, and extract road-level embedding by bipartite masked attention.
%
However, this case does not improve the performance than GMAN since the regional knowledge is not dynamic and limited to provide real-time knowledge. 
%and does not provide how dynamically exploited for real-time.
%
%When we don't have Dynamic convolution, we apply Lz stacks of 5-channel CNN with same padding.
%
When we replace dynamic convolution into 2D-CNN (\#h), the performance gets marginally better than baselines in Table~\ref{tab:result}, but not the best. 
When we do not use gaussian mask on bipartite transform attention, the model takes attention even for distant regions of a road, so it lowers performance (\#i).

\section{Conclusion}

%In this paper, 
%we study how to incorporate real-time regional knowledge to improve road traffic prediction.
%
We propose a novel method that learns spatio-temporal regional knowledge via dynamic convolution with temporal attention and transforms it into the road-level feature via the bipartite transform attention to feed a graph multi-attention for the final output.
We construct three real-world datasets in Seoul consisting of traffic speed and regional population, and evaluation results show that our model outperforms other baseline models in all regions. 
Especially, our model performs significantly better on the roads with more POI as more dynamic regional information is available. 

We plan to investigate how to enable our model to adaptively learn based on the social sense of place by leveraging temporal POI exploitation and land use.   
% \cite{han2020discovering}.
%
We believe our study provides insights for urban planners and researchers who investigate the correlation between the social use of a region and its corresponding traffic.

\bibliography{bibfile1}

\end{document}